\documentclass{article}

\usepackage{microtype}
\usepackage{graphicx}
\usepackage{subfigure}
\usepackage{booktabs} % for professional tables

\usepackage{hyperref}
\usepackage{amsmath}
\usepackage{amssymb}

\usepackage{tikz}
\usetikzlibrary{pgfplots.groupplots,arrows.meta,shadows,positioning,angles,quotes}
\usetikzlibrary{matrix,chains,positioning,decorations.pathreplacing,arrows}
\usetikzlibrary{shapes,arrows}
\usetikzlibrary{arrows,calc,positioning}
\usepackage{xcolor}
\usetikzlibrary{shapes.geometric}
\usetikzlibrary{positioning}
\usepackage{pgfplots}

\usetikzlibrary{external}
\usepackage[accepted]{icml2021}

\icmltitlerunning{Machine Learning Students Overfit to Overfitting}

\begin{document}

\twocolumn[
\icmltitle{Machine Learning Students Overfit to Overfitting}



\icmlsetsymbol{equal}{*}

\begin{icmlauthorlist}
\icmlauthor{Matias Valdenegro-Toro}{rug}
\icmlauthor{Matthia Sabatelli}{rug}
\end{icmlauthorlist}

\icmlaffiliation{rug}{Department of AI, University of Groningen, Groningen, The Netherlands}

\icmlcorrespondingauthor{Matias Valdenegro-Toro}{m.a.valdenegro.toro@rug.nl}

\icmlkeywords{Machine Learning, ICML}

\vskip 0.3in
]



\printAffiliationsAndNotice{}  % leave blank if no need to mention equal contribution

\begin{abstract}
Overfitting and generalization is an important concept in Machine Learning as only models that generalize are interesting for general applications. Yet some students have trouble learning this important concept through lectures and exercises. In this paper we describe common examples of students misunderstanding overfitting, and provide recommendations for possible solutions. We cover student misconceptions about overfitting, about solutions to overfitting, and implementation mistakes that are commonly confused with overfitting issues.  We expect that our paper can contribute to improving student understanding and lectures about this important topic.
\end{abstract}

\section{Introduction}

Machine Learning has had large advances in the last years, fueled by large datasets, large models, and lots of computation. This has driven the demand of machine learning education, particularly about deep neural networks. But the fundamentals of machine learning have not changed, and the basic concept of any successful learning system is generalization, that is, to learn from a limited training set, and still generalize and perform well on test samples that might come from different distributions \citep{hand2006classifier}.

Teaching the concept of overfitting is not easy \citep{demvsar2021hands} \citep{acquaviva2022teaching}, basically because it is a judgment call based on the ratios of training and validation losses. Students learn about overfitting as one of the most basic machine learning concepts, including method to detect it, and regularization techniques to improve generalization and reduce overfitting. Students might incorporate previous experiences into their training \citep{shouman2022experiences}. 

In this paper we show samples coming from our teaching experience in machine/deep learning courses, and show the typical mistakes or misconceptions that students have about the concept of overfitting. We also present samples on misconceptions about how to "solve" overfitting, and discuss the sources of these misconceptions, and propose ideas to improve teaching of both concepts, with the aims of improved learning for this important concept.

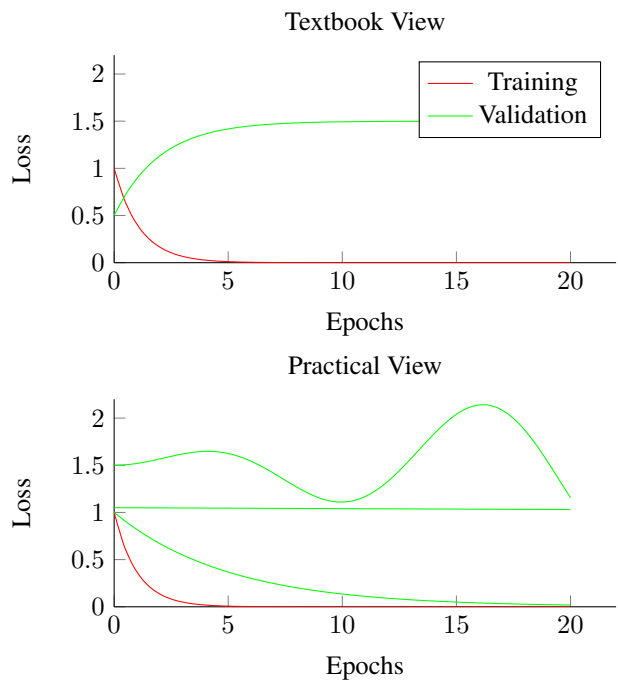
\begin{figure}[t]
    \begin{tikzpicture}
        \begin{axis}[every axis plot post/.append style={mark=none,domain=0:20,samples=50,smooth}, 
            width=\linewidth, height=0.19\textheight, axis x line*=bottom, axis y line*=left, enlargelimits=upper,
            legend pos=north east, ylabel={Loss}, xlabel={Epochs}, ymin=0.0, ymax=2.0, title={Textbook View}] 
            \addplot+[red] {exp(-0.9 * x)};
            \addlegendentry{Training}
            \addplot+[green] {1.5 - exp(-0.5 * x)};
            \addlegendentry{Validation}
        \end{axis}
    \end{tikzpicture}
    \begin{tikzpicture}
        \begin{axis}[every axis plot post/.append style={mark=none,domain=0:20,samples=50,smooth}, 
            width=\linewidth, height=0.19\textheight, axis x line*=bottom, axis y line*=left, enlargelimits=upper,
            legend pos=north west, ylabel={Loss}, xlabel={Epochs}, ymin=0.0, ymax=2.0, title={Practical View}] 
            \addplot+[red] {exp(-1.0 * x)};
            \addplot+[green] {exp(-0.2 * x)};
            \addplot+[green] {exp(-0.001 * x) + 0.05};
            \addplot+[green] {0.04 * sin(9 * pi * x) * x + 1.5};
        \end{axis}
    \end{tikzpicture}
    \caption{Example view of overfitting in textbooks (top) vs how students see overfitting in practice (bottom)}
    \label{overplot}
\end{figure}

The contributions of this paper are a conceptual framework to understand why students have misconceptions on overfitting, we present examples of these misconceptions, both on the concept of overfitting, how overfitting can be prevented, and possible implementation errors that are often be confused with overfitting. We present use cases and recommendations for lecturers to improve teaching of this important concept in machine learning systems.

We believe that this paper can kickstart the discussion about how we teach overfitting to machine learning students, what difficulties they face \citep{marx2022using}, and how innovative methods can improve student understanding, in order for students to become proficient machine learning practitioners.

\section{Concept of Overfitting}

Overfitting is the lack of generalization in a machine learning model \citep{murphy2022probabilistic} \citep{bishop2006pattern} \citep{goodfellow2016deep}. This is usually evaluated over losses computed on train and validation split of the data, where the generalization gap can be estimated:
\begin{equation}
    L_{\text{gap}} = L_{\text{val}} - L_{\text{train}}
\end{equation}
In general if $L_{\text{gap}} >> 0$, it is said that the model is overfitting. But there is normally a small difference between validation and training loss, the question is, how much difference should there be to declare overfitting. Regularization methods like Dropout can also have the effect of inverting the train and validation loss relationship \citep{srivastava2014dropout}. The typical view of overfitting is presented in Figure \ref{overplot}, where training loss decreases with epochs while validation loss increases, clearly indicating overfitting.

Overfitting is generally presented as a binary condition, the model overfits or not. This can be seen in many research papers (some examples are \citep{krizhevsky2012imagenet} and \citep{szegedy2015going}), but overfitting is not formally defined as a binary condition in the literature.

We believe that part of students having misconceptions about overfitting, is that its seen, presented, and possibly taught as a binary condition, and not as a continuous phenomena. Overfitting is more likely to happen with larger $L_{\text{gap}}$ values, and the question is, how large $L_{\text{gap}}$ must be to decide that the model overfits. This is basically a judgment call, and there are no clear guidelines in the literature.

\section{Student Misconceptions of Overfitting}

In this section, we present examples of student misconceptions about the concept of overfitting. We obtained these from our own experience teaching machine learning, deep learning, and reinforcement learning courses at the Bachelor and Master level. Students provided feedback through homework assignments and exams, from where we selected the most common misconceptions specifically about overfitting, with questions directly tasked to evaluate student knowledge about it.

We used the following question to evaluate overfitting concept and possible solutions:
\begin{quotation}
    You are training your favourite neural network and you notice that your validation loss is much higher than your training loss. Give some possible solutions to the problem.
\end{quotation}
Table \ref{table:misconceptions} presents misconception examples for overfitting. The most common misconceptions is to declare overfitting by only looking at the training loss, which by definition does not have the required information to assess generalization, and students also confuse specific loss values with overfitting, like zero training loss, and use non-loss metrics to assess overfitting (commonly accuracy).

Trying to reduce overfitting (we informally call this "solving") is a large part of many machine learning courses, by showing students how to use regularization techniques, reduce model complexity by removing selected layers or decreasing width (number of neurons), or a diverse set of standard model architectures that can be used as a base for their own, in the quest to improve generalization performance. We also note that students have misconceptions on how overfitting can be reduced, examples are presented in Table \ref{table:solving_misconceptions}. Mostly students try to make unrelated changes to the model or training process (like changing learning rates), and in many cases the students do not realize that the dataset is just too small, there are not many chances of learning a useful concept from limited data. One example included tuning the size of the validation or test set by manipulating the splits, which can lead to possible scientific misconduct.

We also identified a set of implementation errors that led students to think their models were overfitting but this was just an incorrect implementation. These are presented in Table \ref{table:impl_errors}, categorized by symptom. Since only losses are typically monitored during training/validation, it is easy to fall prey to implementation errors that produce convincing losses to the untrained eye, while the model is not actually learning.

The biggest implementation error is to test the model with images that are very far from the training distribution. Another important error is not to notice that the model is underfitting, since there is almost no difference between training and validation losses, but this could be caused by not training the model appropriately, or even using incorrect losses or mismatch between model output ranges and labels.

\begin{table}
    \begin{tabular}{p{8cm}}
        \toprule
        A training loss of zero means it is overfitting.\\
        Validation loss is unstable means it is overfitting.\\
        Validation loss that is constant means it is overfitting.\\
        Training and validation loss differ by 0.5 units, my model is surely overfitting.\\
        Validation loss is lower than training loss, means my model is overfitting.\\
        Training accuracy is higher than validation accuracy, my model is overfitting.\\
        Accuracy is constant which indicates the model is overfitting.\\
        Training loss stopped decreasing after $N$ epochs, my model just overfit.\\
        The training loss increases instead of decrease, this means my model is overfitting.\\
        My model overfits from the start of training.\\
    	Overfitting only happens within Supervised Learning.\\
        I have 10K data points, my model cannot possibly be overfitting.\\
        My network is pre-trained on ImageNet, thus it cannot overfit. \\
	\bottomrule
    \end{tabular}
    \caption{List of Student Misconceptions on the concept of Overfitting, ordered by approximate frequency (most frequent on top).}
    \label{table:misconceptions}
\end{table}

Another important source of confusion regarding failure to generalize, happens with the test distribution. Students might train a model on the MNIST dataset, and then download digit images from the internet or draw their own in painting software, and then test the model on these images. A very likely result is that the model will fail to make correct predictions, and this is simply because the new test distribution is very far from the training distribution, only MNIST-like digits have a chance of actually working on this model.

Students generally learn overfitting using supervised learning examples, but it can also happen in unsupervised and reinforcement learning settings

\section{Beyond Supervised Learning}

As the concept of overfitting is typically explained from the Supervised Learning perspective, students are prone to forget that this phenomenon might appear in other machine learning paradigms as well. As a result they tend to overlook the existence of overfitting within Unsupervised Learning (UL) and Reinforcement Learning (RL) and fail in realizing that if their models underperform, overfitting might be the cause of their underwhelming results.  
	 
\begin{itemize}
	\item\textbf{Unsupervised Learning}. A common UL algorithm which is taught in most undergrad courses is \texttt{K-Means}. It is well known that the algorithm requires the user to specify in advance the number of $k$ clusters, which are then found through an iterative process that minimizes the within-cluster variance. One easy way that students find to ensure that such quantity is minimized is to set the value of $k$ as high as possible. Yet, they tend to forget that by doing so they will obtain clusters that are only specific to very few data-points, therefore failing in partitioning the data space properly.

	\item \textbf{Reinforcement Learning}. Just like a classifier might fail in generalizing to data samples that are outside from the training distribution, so can a RL agent that has to deal with states that have not yet been seen throughout its interaction with the environment. This issue is even more prominent nowadays as RL algorithms make use of Experience Replay memory buffers that store RL trajectories that are needed for training, therefore resulting in agents that can only learn a value-function or policy for a limited set of trajectories.   
    
\end{itemize}

In many cases methods for UL and RL are reformulated as supervised learning problems, for example autoencoders for UL, and Deep Q-Networks for RL.

\begin{table}
    \begin{tabular}{p{8cm}}
        \toprule
        If my model overfits, I can solve it by decreasing the learning rate.\\
        Overfitting can be solved by adding more convolutional layers.\\
        My model overfits because I did not normalize the inputs correctly.\\
        I am using ResNet (or any other standard model) and this model cannot overfit, there must be a bug somewhere.\\        
        I will change the train/validation split from 80/20 to 70/30 and this will help with overfitting.\\	    
        Overfitting is solved by changing some hyper-parameters (unrelated to regularization or model complexity).\\
        \bottomrule
    \end{tabular}
    \caption{List of Student Misconceptions on "solving" Overfitting}
    \label{table:solving_misconceptions}
\end{table}

\begin{table*}
    \centering
    \begin{tabular}{p{6.5cm}p{10cm}}
        \toprule
        Issue or Symptoms								& Possible Causes\\
        \midrule
        Model predicts incorrectly on new images 		& Input normalization not applied to new images.\\
                                                        & Poor out of distribution generalization.\\
                                                        & Inputs are very different from the training set.\\
        \midrule                                                                        
        Training and validation loss are both high.		& Model was not trained for enough epochs to reach convergence.\\
                                                        & Model is underfitting / model is not appropriate for the task.\\
                                                        & Incorrect training loss (e.g. using classification loss for regression).\\
                                                        & Incorrect activation at the output or mismatch in output ranges.\\
                                                        & Mismatch between model task and dataset task.\\
        \midrule
        One or both losses behave erratically (noisy)  	& Not enough training or validation data.\\
                                                        & Model is too overparametrized for the training set.\\
        \midrule
        Other and miscellaneous							& Metrics are used that are not appropriate for the task.\\
                                                        & Losses are not differentiable or implemented incorrectly.\\
        \bottomrule
    \end{tabular}
    \caption{List of Implementation Errors that can be confused with Overfitting}
    \label{table:impl_errors}
\end{table*}

\section{Use Cases for Learning about Overfitting}

We believe that multiple definitions of overfitting can be used to show that it is not a unique concept, and there are multiple aspects to be considered. For this we suggest the following exercises or use cases that can aid students to build a deeper understanding of overfitting.

\begin{description}
    \item[Multiple Test Sets] Train a model on a standard dataset, that contains its own test split, but provide additional test sets, including images collected from the internet, and own images. Students should see that the model works on the standard test set, but might fail on new images from the internet. This can be used to show that any model can overfit, only depending on the data it is tested on.
    \item[Multiple Tasks] Select multiple tasks (not just classification), to showcase different scenarios where a model can overfit. For example be object detection, semantic segmentation, or even instance segmentation (a multi-task combination of object detection and segmentation). These tasks have different sample complexity and data requirements.
    \item[Vary Training Set Size] Overfitting often happens when there is not enough training data, and providing students with sub-sampled training sets to force overfitting can help them understand the concept, specially if this is combined with a variable size training set. An exercise can be built, adding more data to the training set, the student can see how overfitting decreases and generalization improves. This particular use case can be used to showcase that overfitting is not a binary condition but has continuous properties.
    \item[Multiple Paradigms] We have argued that students might think that overfitting is specific to supervised learning. Using RL or UL tasks and datasets might help dispel this misconception. For example, using unsupervised feature learning on a standard dataset, students can explore different classes present on the feature learning dataset, then using a different dataset that might or might not share classes with the feature learning dataset, which would reveal the different levels of generalization on seen and unseen object classes. In reinforcement learning, limited exploration can be used to show students the influence of the training set on generalization, using part of the environment for training, and out of training distribution parts of the environment for testing, which will most likely reveal failure to generalize.
    \item[Showcase Judgement Calls] We argue that the generalization gap $ L_{\text{gap}}$ is not intuitive to interpret since there are no clear thresholds to declare overfitting. All previous examples can be used to show this to students, where different generalization gaps can be produced by the teacher, with different overfitting results.
\end{description}

The overall concept in these use cases is to teach overfitting separately from failure to generalize. Overfitting should be taught with multiple use cases and datasets, particularly test sets, not just on a single task/dataset or the standard train/validation split plot like Figure \ref{overplot}.

In the appendix we provide a checklist that students and lecturers can use to check if their training scheme is appropriate. This can be useful to systematically debug overfitting and failure to generalize issues, and from where new use cases or exercises can be derived \citep{raschka2022deeper}.

We encourage that lecturers to make "negative" examples, where there is sample code but with some implementation errors, and students are tasked to correct these errors (for example, wrong losses, wrong number of output neurons, output range mismatch, etc), and improve their knowledge about overfitting.

\section{Conclusions and Future Work}

In this paper we study how to teach overfitting through presenting examples of student misconceptions about the concept, how it can be "solved", and implementation errors that can be confused with overfitting. We present possible use cases and exercises that we believe can help students improve their understanding of overfitting in practice, and move away from a binary condition into a continuous phenomena.

We believe that this is useful information to improve teaching of the overfitting concept and reduce possible student failure. We encourage lecturers and practitioners of Machine Learning to create exercises, homework, and assignments specifically crafted to address these misconceptions, as a way to put learning materials ahead of possible student understanding, which should overall improve learning outcomes.

As future work we plan to develop special exercises to tackle the overfitting misconceptions, and evaluate them in future versions of our courses on a controlled experiment.


\newpage
\bibliography{biblio}
\bibliographystyle{icml2021}

\newpage
\appendix
\onecolumn

\section{Checklist for Debugging ML Models}

Here we provide a short checklist for students to debug the training of their machine learning models. The idea of this checklist is to follow it step by step

\begin{enumerate}
    \item \textbf{Is the loss used for training the model appropriate for the task?}\\
          If the incorrect loss function is used (cross-entropy for classification, mean squared error for regression, or similar), then the model might not train, the loss might not decrease, or the model might not converge at all.
    \item \textbf{Was the model trained until convergence?}\\
          A common source of confusion is looking at metrics/losses or predictions of a model that has not been trained for a sufficient amount of epochs. The predictions might look random or with strange patterns, which is produced by the partial and unfinished training. Looking at predictions only makes sense once the model has converged, with a loss that does not decrease further after a long stretch of constant decrease.
    \item \textbf{Was validation data used to be able to check for overfitting?}\\
          There is only one way to check for overfitting, that is, to use a train and a validation split that have no samples in common, train the model on the training set, and after each epoch (or a set number of iterations), evaluate the model on the validation set. Losses should be computed on the train and validation predictions, and compared to decide if overfitting can happen.\\
          The variation of train and validation losses can be plotted (see Figure \ref{overplot}) to decide if overfitting is happening, with a gap ($L_{\text{gap}}$, of varying size) indicating overfitting, specially if this gap grows with increasing number of epochs.\\
          Note that metrics should not be used to assess overfitting, since a metric might measure different aspects of the models' predictions than the loss. Only train and validation losses should be used to assess overfitting.
    \item \textbf{Is there enough training data?}\\
          Assess the number of samples of your training set. If its too small, then the model might not learn the desired pattern or have unstable loss and convergence properties. The model might fail to generalize, since there is little information available for learning. Compare the number of samples in the training set to the number of model parameter, and assess if the system is well specified or underspecified (the latter would be the case if there are more model parameters than training samples). If this is the case, consider acquiring more data or using data augmentation.
    \item \textbf{Is the data distribution of training and validation/test sets equal or similar?}\\
          If input data used for testing is very different than the training set, the the model might not generalize to these images, and this is a failure to generalize, not exactly an overfitting problem.
    \item \textbf{Is the model and/or training process correctly implemented?}\\
          Assess if there are implementation errors (refer to Table \ref{table:impl_errors}) that could cause the model to fail training, which might masquerade as overfitting. Check if losses are implemented correctly, labels are correctly loaded and have the expected ranges, check if gradients have the expected values (to assess vanishing gradients), etc.
\end{enumerate}

\end{document}